\documentclass[letterpaper, 10 pt, conference]{ieeeconf}  

\IEEEoverridecommandlockouts                              

\usepackage{bbm}
\usepackage{mathtools}
\usepackage{amssymb}
\usepackage{graphicx}
\usepackage{multirow}
\usepackage{color}

\usepackage{hhline}
\usepackage{xspace}

\newcommand{\ie}{\emph{i.e.}\xspace}

\newcommand{\etc}{\emph{etc.}\xspace}

\title{\LARGE \bf
Adaptive Deep Learning through Visual Domain Localization
}

\author{Gabriele Angeletti$^{1}$ and Barbara Caputo$^{1}$ and Tatiana Tommasi$^{1}$
\thanks{*This work was supported by the ERC grant 637076 - RoboExNovo.}
\thanks{$^{1}$ G. Angeletti, B. Caputo, T. Tommasi are with the VANDAL Laboratory, Department of Computer, 
Management and Control Engineering (DIAG), Sapienza Rome University, Rome, Italy
and with the Italian Institute of Technology, Milan, Italy
        {\tt\small tatiana.tommasi@iit.it}}%
}

\begin{document}

\maketitle
\thispagestyle{empty}
\pagestyle{empty}

\begin{abstract}
A commercial robot, trained by its manufacturer to recognize a predefined number and type of objects, might be used 
in many settings, that will in general differ in their illumination conditions, background, type and  degree of clutter, and so on.
Recent computer vision works tackle this generalization issue through domain adaptation methods, 
assuming as source the visual domain where the system is trained and as target the domain of deployment. 
All approaches assume to have access to images from all classes of the target during training, 
an unrealistic condition in robotics applications. 
We address this issue proposing an algorithm that takes into account the specific needs of 
robot vision. Our intuition is that the nature of the domain shift experienced mostly in robotics
is local. We exploit this through the learning of maps that spatially ground the domain 
and quantify the degree of shift, embedded into an end-to-end deep domain adaptation architecture. 
By explicitly localizing the roots of the domain shift we significantly reduce the number of parameters of the 
architecture to tune, we gain the flexibility necessary to deal with subset of categories in the target domain at 
training time, and we provide a clear feedback on the rationale behind any classification decision, which can be 
exploited in human-robot interactions. Experiments on two different settings of the iCub World database confirm 
the suitability of our method for robot vision. 
\end{abstract}

\section{Introduction}
\label{sec:intro}
The importance of vision for robotics is pervasive: from self-driving cars to detecting and handling 
objects for service robots in homes, from kitting in industrial workshops, to robots filling shelves 
and shopping baskets in supermarkets, \etc. All these applications imply interacting 
with a wide variety of objects and being able to recognize them in unconstrained environments. To this end, robots 
must deal with significant changes in the visual appearance of objects when being deployed in various \emph{visual domains}.
Objects may look very differently due to changes in illumination, 
background, scale, rotation and 
occlusion, when for instance a portion of the object is outside the 
viewing area. Thanks to the recent adoption of high capacity deep neural networks for many robotics vision 
tasks, the effect of domain shift due to the described variations have been alleviated but not removed
and the field is in need of reliable domain adaptation (DA) methods. 

We focus on this problem, starting from the intuition that at least some of the visual characteristics 
of each domain can be localized in the images and this spatial grounding \cite{roots} can guide 
a tailored domain adaptation strategy.
Differently from existing CNN-based domain adaptation methods that try to minimize the discrepancy 
between source and target domains while learning an object classifier on the source (we refer to section 
\ref{sec:related} for a review of previous work), we separate the 
two objectives. First we study how to localize the image regions more responsible for the domain 
shift as well as the regions that are shared among the domains. Then we use the obtained map to 
guide the attention of the source classifier. This path leads to a modular solution that is
particularly suitable for robotics applications under several aspects:\vspace{1mm}
\begin{itemize}
\item the information pre-saved in the maps allow to lighten the subsequent classification learning
procedure by using pooling layers with adaptive window size that reduce the number of network 
parameters compared to standard architectures;\vspace{1mm}
\item the maps allow to get the rationale at the basis of the adaptive procedure. This is particularly
relevant for human-robot interaction because the agent will be able to provide visual explanations supporting 
its final decision;\vspace{1mm}
\item the method can easily deal with classification problems in the target domain that differs from those
experienced by the robot during training, because the maps may be learned on a subset of the target classes.
This is extremely important for robot applications, where the details of the deployment task is not known 
a priori and it is unlikely to have transductive access to the whole target data.

\end{itemize}
We name our method \emph{Local Adaptive} (LoAd) deep visual learning and we show that, besides
providing a deeper understanding of the domain shift problem for robotics applications, it also 
gets better results than competing state of the art domain adaptation methods 
on two robotics settings.

The rest of the paper is organized as follows: section \ref{sec:related} briefly reviews the state
of the art in domain adaptation and CNN visual analysis, section  \ref{sec:load} describes the LoAd 
network and section \ref{sec:exper} presents the experimental evaluation with a discussion on the
role of each network component to the final cross-domain object classification performance. Section
\ref{sec:concl} offers our final remarks and indicates possible directions for future research.

\section{Related Work}
\label{sec:related}
\subsubsection{Domain Adaptation}
The goal of domain adaptation is to produce good models on a target domain, by training
on labeled data from the source domain and leveraging unlabeled samples from the target domain
as auxiliary information during training. The problem of reducing the domain shift
between train and test data was first tackled in the area of natural language processing
\cite{domain-adaptation-review} and in the last years has gained an increasing attention in computer vision 
for solving dataset bias in object recognition and detection \cite{DBLP:journals/corr/Csurka17}.
Although much more robust than previous learning technologies, deep learning is still
affected by domain shift \cite{decaf} in particular in the considered unsupervised 
setting where no labeled target data are available and standard fine-tuning is not applicable \cite{finetune}.

We can identify two main solving directions, one based on \emph{instance re-weighting}
and the other on \emph{feature alignment}. 
In the first case, the basic approach consists in  
evaluating the similarity of source instances to the target with the aim of 
balancing their importance or eventually sub-select them before learning a model.
Different measures of similarity have been proposed in combination with shallow 
learning methods for this weighting procedure \cite{KMM,landmarks,ChuDC13}.
More recently in \cite{Zeng2014} a deep autoencoder 
was trained to weigh the importance of source samples 
by fitting the marginal distributions of target samples for pedestrian detection.
The second adaptive solution based on feature transformation and alignment has been 
declined in a large number of ways, all based on searching a common 
subspace to minimize the difference among the corresponding domain
distributions. Feature transformation was obtained through metric learning in \cite{office,Kulis2011},
PCA in \cite{Fernando2013b}, while multiple intermediate projection steps were considered in \cite{GongSSG12}. 
Intermediate features were also obtained in \cite{coral} by aligning source and target covariance matrices. 
Deep learning architectures for object classification have been modified to 
accommodate the second objective of minimizing a domain divergence measure \cite{Long:2015,dcoral,residual}. 
An alternative way to measure domain similarity is that of 
discriminating among them and using the domain recognition loss
in an adversarial min-max game while training for object classification \cite{DANN}.
Finally, a different solution based on the introduction of adaptive network layers for 
tunable batch normalization has shown high performance on several object classification 
benchmark datasets \cite{carlucci2017auto}.
All the existing CNN-DA methods, with the notable exception of \cite{openset2017}, are restricted by the assumption that
samples of source and target share the same label space and that data of all the categories are provided
at training time. In case only a subset of the target categories is available during training, a 
possible solution is to restrict also the source category set, but this 
implies that the classifier will not cover all the classes and it should be trained again in
case new unlabeled target samples of the remaining classes become available. Alternatively
the whole source can be used for adaptation with the partial target but this unbalanced condition
generally affect the final classification performance. Our work overcomes this limitation, by proposing a modular 
approach that yields robustness with respect to variations in the number of classes from the source to the target domains.  

In the latest robotics literature a lot of attention has been dedicated to adaptive methods for
agents deployed in the real world but trained in simulation environments and with synthetic data 
produced or collected from free Web resources \cite{sim2real,abbeelIROS17, tzengWAFR16}. 
Although bridging the so-called reality-gap
is important to allow a reduction in the need of manually annotated data, robotic
perception provides ample motivations for exploring domain adaptation methods even within real 
world settings, when changing the visual conditions \cite{WulfmeierIROS2017} or when
transferring the knowledge acquired by one robot to another \cite{acrossRobots}.
In our work we focus on a single robot that needs to recognize object categories undergoing
significant appearance changes due to scaling and translation and we show that the corresponding
domain gap can be reduced with a tailored localized adaptive solution based on the identification of
domain-invariant image regions.

\vspace{2mm}
\subsubsection{CNN Visual Analysis} One of the advantages of convolutional neural networks
is that they can be made transparent by visualizing the region of the input images that are
important for the trained model. 
A widely adopted strategy to get those visualizations consists in starting from 
individual feature maps at any layer in the network and project them back towards the input image 
to identify the patches more responsible for strong activation \cite{Simonyan_ICLR_2014,guidedbackprop}. 
The produced heatmaps are informative at high level but often they are not class discriminative.
A recent method named \emph{Class Activation Mapping} \cite{zhou2015cnnlocalization} replaces
fully-connected layers of image classification CNN architectures with convolutional layers and 
global average pooling to obtain class-specific feature maps, but  
the network modifications tend to reduce the accuracy performance in the final classification task. 
The \emph{Grad-CAM} approach introduced in \cite{gradcam} does not need any change in the network architecture: 
on the basis of class specific gradients, it re-weights and combines the convolutional feature maps.
This method produces class-discriminative localization heatmaps as good as those obtained
with strategies that directly start from the input images and either use a mask-out procedure occluding
specific object parts, or classify multiple image patches to collect class-wise scores per pixels \cite{zeiler13,Oquab:2014}. 
However, differently for these last approaches which needs several pass per image through the network,  
Grad-CAM has a single forward and partial back-ward pass per image, resulting an order of
magnitude more efficient.

Recently patch-occlusion CNN visualization has been used to search for the spatial roots of visual domain 
adaptation \cite{roots} and produced the so called \emph{domainness maps} that indicate the localization of 
domain-specific and domain-generic regions. Features at different levels of domain specificity were extracted
and evaluated for cross-domain object classification with shallow models. In our work we propose a different 
and more efficient strategy for domain localization by following \cite{gradcam} and we integrate the obtained
domainness maps into an end-to-end deep learning architecture.
\begin{figure*}[tb]
\centering
\includegraphics[width=0.8\textwidth]{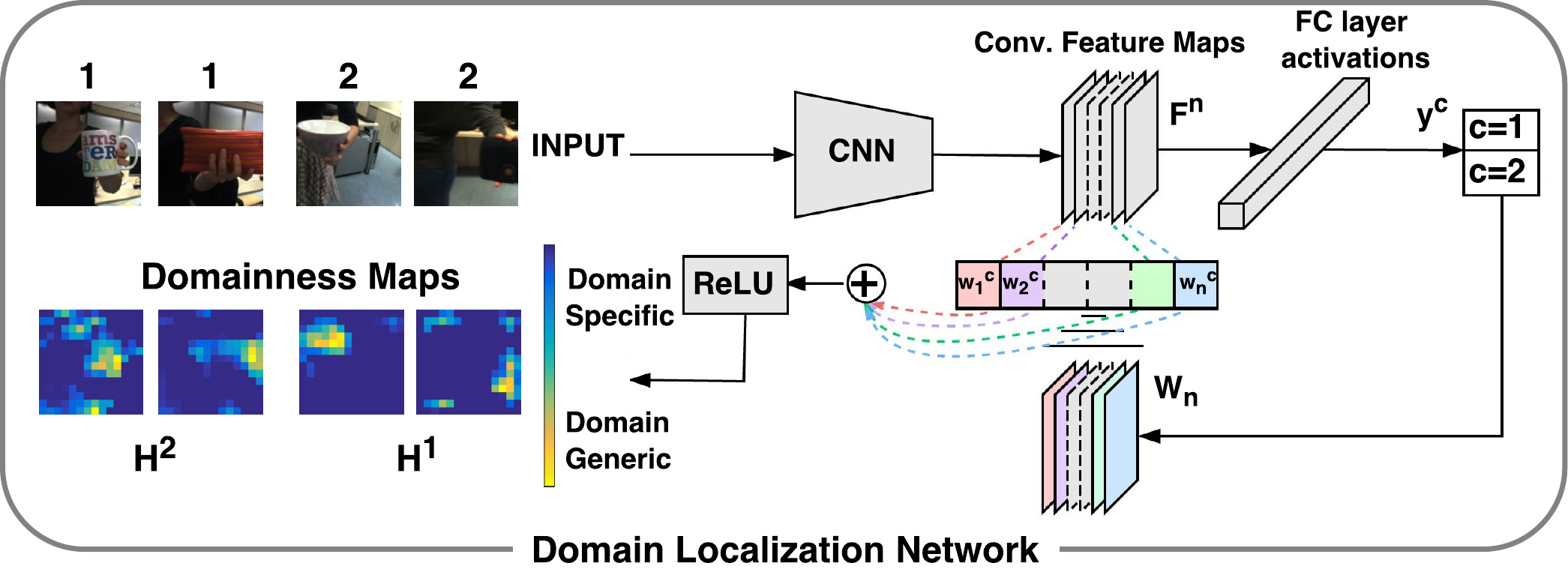}
\caption{Outline of the Domain Localization Network that produces the domainness maps.
Here we consider as example the domain shift induced by object translation.  
As it  can be expected, domain-generic regions are localized in correspondence of the objects
while the background is domain specific. We use the same notation as in sec. \ref{sec:load},
we get a set of $W_n$ filters per class that will guide the training of LoAd (see Fig. \ref{fig:load}). Figure adapted from \cite{gradcam}.}
\label{fig:gradcam}
\end{figure*}

\section{LoAd: Local Adaptive Network}
\label{sec:load}
\begin{figure*}[tb]
\centering
\includegraphics[width=0.95\textwidth]{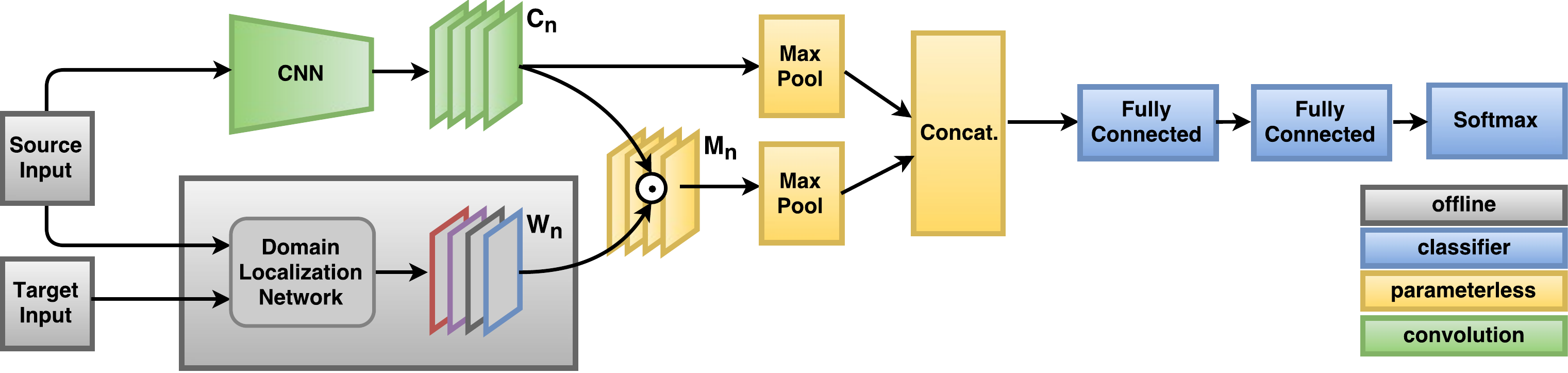}
\caption{An illustration of our Localized Adaptive (LoAd) Network. The colors
are used to code different part of the architecture. With yellow we indicate that the multiplicative 
and concatenation layers execute operations that do not need parameter learning. 
The layers that are instead directly involved in the network parameter learning procedure are 
the green and the blue ones, corresponding to the convolutional and fully connected layers. 
Note that the final softmax performs object classification
on the annotated source data, while the pre-trained domain localization network needs as input both 
the source and the target data but is unsupervised with respect to the object category labels.}
\label{fig:load}
\vspace{-2mm}
\end{figure*}

The intuition at the basis of our Local Adaptive (LoAd) network is that, by discovering
the image regions which contain information shared by the two domains, it will be possible to 
highlight them when learning the source classifier and obtain a model robust to domain shift.
While designing the network, we need to answer two questions: how to localize domain-specific 
and domain-generic image regions, and how to integrate the obtained information to guide the 
attention of a deep learning model trained for object classification. 
The outline of the network and technical details for both answers are described in the 
following.

\vspace{2mm}
\subsubsection{Domain localization}
\label{subsec:domloc}
Discriminative information about two domains can be easily collected by training a deep model
that differentiates among them. These information are stored in the inner layers of the network:
inspired by \cite{gradcam}, we exploit the gradients flowing into the 
final convolutional layer of a CNN to produce coarse maps about the spatial 
grounding of the domain in an image.

Basically the source and target images are given as input to the network sketched in Figure \ref{fig:gradcam},
which is pre-trained on Imagenet and fine-tuned on a single sigmoid unit for binary classification.
We indicate with $y^c = f(x)$ the softmax score at the last layer. Here $f()$ is the input-output 
network mapping, and $c=1,2$ are the classes \ie the domains in our case. The last convolutional layer 
produces $n = 1, \ldots, N$ feature maps $F_{ij}^n \in \mathbb{R}^{u \times v}$, where $i,j$ span on 
the width $u$ and height $v$.
These feature maps retain spatial information and we can measure how much each of them 
contributes to the final score for a class by calculating the derivatives $\frac{\partial y^c}{\partial F_{ij}^n}$
which are then global average pooled in order to get a single number: 
\begin{equation}
w_n^c = \overbrace{\frac{1}{Z}\sum_{i=1}^u\sum_{j=1}^v}^\text{global average pooling} \underbrace{\frac{\partial y^c}{\partial F_{ij}^n}}_\text{gradients via backprop}~.
\end{equation}
We can see that the derivative is positive if an increase in the value of the pixel $F_{ij}^n$ yields an increase of 
the value of $y^c$. These weights are then used to define a rectified weighted linear combination of the
feature maps 
\begin{equation}
H^c = ReLU\left(\sum_{n=1}^N w_n^c F^n\right) \in \mathbb{R}^{u \times v}
\end{equation}
which can be seen as coarse heatmaps. When starting from the AlexNet \cite{alexnet} 
basic architecture we have $N=256$ 
and $u = v = 13$. 
For visualization the heatmaps can be upsampled through bi-linear interpolation and we
can use them to identify image regions shared by the domains, or specific of each of them: given an image 
of domain $c=1$, the regions highlighted as important for domain $c=2$ correspond to the \emph{domain-generic} 
areas, while the regions highlighted as important for domain $c=1$ are \emph{domain-specific}. 
Same holds when inverting the domains. 
By following \cite{roots} we indicate the obtained heatmaps as \emph{domainness maps}.
Together with the network architecture, Figure \ref{fig:gradcam} shows 
examples of domainness maps which highlight domain generic image regions.

\vspace{2mm}
\subsubsection{Spatial attention by multiplicative fusion}
Besides providing visual explanations about \emph{where} the domain insists in an image,
the procedure described above can be used to reduce the domain shift by guiding a learning 
procedure to attend the regions shared by the two domains. With this goal in mind we keep
the \emph{per-feature map activations} from the convolutional layer without summing over depth
\begin{equation}
W^c_n = ReLU\left(w_n^c F^n\right) \in \mathbb{R}^{u \times v}\quad\quad\text{for } n = 1\ldots N,
\label{eq:mapactivation}
\end{equation}
in this way the information about how each filter contributed to the decision of a certain
class (\ie domain) is integrated in the filter itself. 
We propose to exploit the knowledge provided by these filters through \emph{multiplicative fusion} 
\cite{multfusion} when training a network for object classification on the source.

A deep network with the same basic structure of that used for domain recognition (AlexNet) 
pre-trained on Imagenet and fine-tuned to minimize a softmax loss on the source object categories,
will produce feature maps $C$ of the same size of $W$ at the last convolutional layer. 
Object and domain knowledge are then integrated by element-wise multiplication 
$M = C \odot W$ to get convolutional maps that attend only domain-shared regions.
The ReLU function in (\ref{eq:mapactivation}) is strictly selective and may reduce to zero
many pixels also in gray areas of the image filters where domain specificity is questionable.
With the aim of enhancing the domain-shared information without loosing information
from the overall image appearance, we add back the original object filters by concatenating $C$ to
$M$\footnote{We also tried the alternative solution of using only multiplicative fusion in 
the form $M = C \odot (W + 1)$ with lower results.}.

\vspace{2mm}
\subsubsection{LoAd Implementation}
The final architecture of our LoAd network can be seen in Figure \ref{fig:load}. From the input layer to the last
convolutional layer, the architecture is the same as a standard AlexNet.
Then the output feature maps
are duplicated: one copy is kept as in its original form, the other is multiplicatively fused with the
domain localization filters. Max pooling with an adaptive choice of the kernel and
stride dimensions (see Section \ref{subsec:ablation}) is performed on each of the two branches which are then recombined 
by concatenation producing enriched feature maps with a highlight on the domain generic part of the images.
The network ends with three fully connected layers with dimension $2048\rightarrow 2048 \rightarrow K$, where
$K$ is the number of object categories to be recognized through the minimization of a softmax loss.

We implemented the LoAd network\footnote{https://github.com/blackecho/LoAd-Network}
with Torch7 \cite{torch7}. 
Element-wise multiplication as well as concatenation are basic operations in Torch and 
could be easily used to prepare the filter integration layers. The adaptive max pooling layer was 
implemented as an extension of Torch7 while the domain localization network was implemented by starting
from the code of \cite{gradcam} and we modified it according to the description 
provided above. 
We always start from Imagenet pre-trained models, so for both networks the initial layers
till the last fully convolutional one are frozen while the remaining layers are either fine-tuned
or trained from random initialization with learning rate  $5\times 10^{-4}$  and 
weight decay of $5\times 10^{-4}$ as regularizer. For the fully connected layers the regularization is
obtained through a dropout of 0.6. All these hyperparameters have been chosen using a validation set.
The optimizer used in the training procedure is Stochastic Gradient Descent with momentum 
parameter \cite{momentum} set to 0.9  and  Nesterov method enabled \cite{nesterov-momentum}. We used a batch 
size of 256 and training was performed in parallel on four NVIDIA Titan X GPUs.

\section{Experiments}
\label{sec:exper}
\begin{figure*}[tb]
\includegraphics[width=1\textwidth]{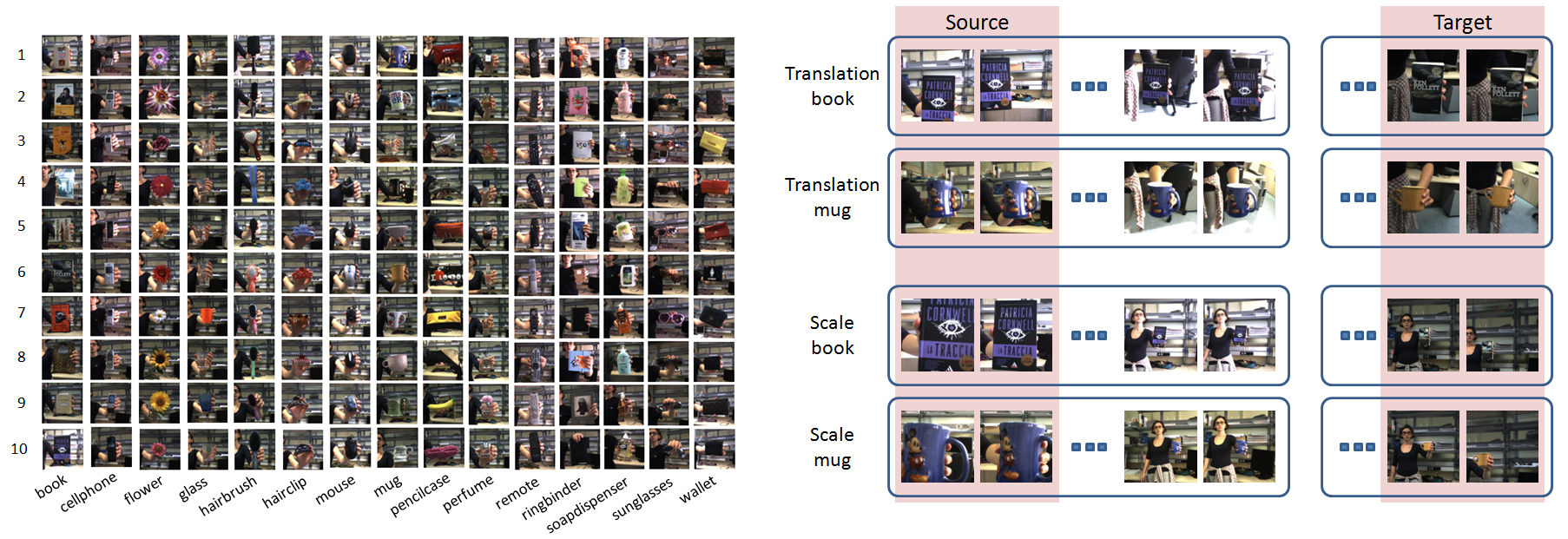}
\caption{\textbf{Left}: Overview of the 15 object categories of the \emph{iCubWorld Transformation} dataset with 10 instances per category. Figure adapted from \cite{Pasquale2016IROS}. \textbf{Right}: Illustrative 
description of the two domain adaptation tasks designed for our experiments. 
Out of the whole set of translation and rotation images, only the first 50 samples are considered
as source data, while the last 50 samples define the target. To simplify the figure we show only 2 out of the
beginning and ending 50 images and only 2 objects (book, mug) out of the whole set of 15 categories.
Often the objects are not centered or can be seen only in part due to the large scale. 
Note also that the the set of object instances in the source and in the target do not overlap.}
\label{fig:dataset_setting}
\vspace{-2mm}
\end{figure*}

In this section we describe the experimental setup used to evaluate our LoAd network 
(\ref{subsec:setup}) and the produced experimental results (\ref{subsec:analysis}). 
We also provide a detailed ablation study (\ref{subsec:ablation}) on different 
components of the proposed approach. 

\subsection{Experimental Setup}
\label{subsec:setup}
For our experiments we use the \emph{iCubWorld} dataset which was created as a benchmark 
for object recognition in robotics \cite{Pasquale2015MLIS}.
A human teacher shows a set of objects to the iCub robot, which uses a tracking routine to follow 
them with its gaze. Supervision is in form of objects' labels provided verbally by the human. In particular we
focus on the \emph{iCubWorld Transformations} \cite{Pasquale2016IROS} (see Figure 
\ref{fig:dataset_setting}-Left) which comprises 
150 objects evenly divided into 15 categories with their appearance changing: each object is acquired 
while undergoing isolated visual transformations in order to study invariance to real-world nuisances.
We considered two cases (see Figure \ref{fig:dataset_setting}-Right):
\vspace{2mm}
\begin{list}{\labelitemi}{\leftmargin=1em}
\item{\textbf{Translation:}} the human moves in semi-circle around the iCub robot keeping approximately the same
distance and pose of the object in the hand with respect to the cameras. Out of the whole rotation, that
is covered on average by 150 images, we define two domains by considering only the first and last 
50 images showing the object at the extreme of the semi-circle, starting on the left and ending 
on the right. Indeed in this movement the appearance of the object background changes significantly 
as well as the illumination conditions, while the object remains the same.
\vspace{2mm}
\item{\textbf{Scale:}} the human moves the hand holding the object back and forth, thus changing the object scale with
respect to the cameras. Also in this case the whole movement is depicted in 150 images on average and 
we pick only the first and the last 50 images to define our domains. The object in its extreme far and 
extreme close positions to the camera occupies a different portion of the image, thus inducing
a significant changing in its overall appearance.
\end{list}
\vspace{2mm}
Since every category contains 10 object instances we divided them into 6 and 4 for the two domains,
introducing a small unbalance that can be naturally present between source and target. Thus in both 
transformation cases, one domain (\emph{left}~/~\emph{close}) contains 4500 images while the other
(\emph{right}~/~\emph{far}) contains 3000 images. Note also that the the set of object instances in the 
source and in the target do not overlap. Both domains are used as source and target in
turn in our experiments. In particular we consider two settings: 
\vspace{2mm}
\begin{list}{\labelitemi}{\leftmargin=1em}
\item{\textbf{Adapt on whole-target}:} the whole labeled source data and the whole unlabeled target
data are exploited during training and adaptation. At test time the learned model is used to annotate
the target samples. 
\vspace{2mm}
\item{\textbf{Adapt on sub-target}:} during training all the source data are available, but
only a sub-part of the target is provided. Specifically, the source is composed by samples of all the
15 object categories, while the target visible at the training phase contains only 8 object categories.
Thus, while the source classifier can still be trained to recognize 15 classes, the joint adaptation process
can leverage only 8 object categories. At test time the whole 15 object target set should be annotated, 
containing both the classes available during training and the initially unseen 7 categories.
\end{list}
\vspace{2mm}

To quantitatively verify the presence of the visual domain shift between the described data domains
we run a first set of experiments by training a classifier on the source and then comparing the performance
while testing on the source and on the target \cite{office}. We started by extracting features
from all the images by using the second fully-connected layer (fc7) of AlexNet 
pretrained on Imagenet which provides a representation vector of 4096 dimensions for each image. 
The source domain is then randomly divided into 80\% - 20\% sets respectively used for training and testing a linear SVM classifier. 
The same model is finally tested also on the target. The obtained results are reported in 
Table \ref{table:domain_shift} and show a drop in performance  which indicate a significant amount of domain shift,
even more evident in the scale case than in the translation one.

\begin{table}[tb!]
\caption{Evaluation of the domain shift in our experimental setting 
when representing the images with AlexNet fc7 features. Here S/T stand 
for source/target, while X $\rightarrow$ Y means that the classifier 
is trained on X and tested on Y. The presence of domain shift is indicated
by the large drop in performance between S $\rightarrow$ S and S $\rightarrow$ T.}
	\begin{center}
		\begin{tabular}{|c|c|c|c|c|} 
                  \hline 
         &        S 				&  			T			& {S $\rightarrow$ S} & {S $\rightarrow T$} \\
                  \hline
    {\multirow{2}{*}{translation}}     &         left 	& right &98.33 & 45.80 	 \\
         								&         right & left  &99.33 & 54.49	\\ \hline
    {\multirow{2}{*}{scale}}    &         close & far	&99.45 & 18.44	 \\
         						&         far  	& close &98.67 & 28.80	\\
                  \hline
        \end{tabular}
	\end{center}
	\label{table:domain_shift}
    \vspace{-4mm}
\end{table}

\begin{table*}[tb!]
\caption{Percentage classification accuracy of our LoAd network and several baselines on the defined iCubWorld Transformation
settings. To take into consideration possible small fluctuations due to the network batch learning, the experiments 
with deep methods have been repeated five times and we report here the average results with their standard deviation. The best results for 
each setting are highlighted in bold.}
	\begin{center}
		\begin{tabular}{|c|c|c|c|c||c|c|c|} 
        \hline  
\multicolumn{2}{|c|}{}& \multicolumn{3}{c||}{iCubWorld Translation} & \multicolumn{3}{c|}{iCubWorld Scale} \\
        \cline{3-8}
\multicolumn{2}{|c|}{}& left $\rightarrow$ right 	& right $\rightarrow$ left 	& average 	& close $\rightarrow$ far 	& far $\rightarrow$ close & average\\
        \hline      
 
\multicolumn{2}{|c|}{Alexnet}		& 50.41 $\pm$ 0.98 			& 54.01 $\pm$ 0.59		& 52.21		& 18.20 $\pm$ 0.71	& 27.45 $\pm$ 1.23		& 22.83 \\
\hline
				& {DANN}	\cite{DANN}	& 62.60 $\pm$ 1.29		& 41.08	$\pm$ 0.10	& 51.08		& \textbf{38.86	$\pm$ 1.86}	& \textbf{47.87	$\pm$ 0.31}	&	\textbf{43.37}	\\
adap on 		& {Auto-DIAL}	\cite{carlucci2017auto}& 58.51 $\pm$ 1.16		& 60.67 $\pm$ 1.21	& 59.59 	& 36.33	$\pm$ 0.10   & 42.58 $\pm$ 0.12&	39.46	\\
whole-target	& {ROOTS}  \cite{roots}		& 54.29					& 55.29							& 54.79		& 20.07				& 34.71			&	27.39	\\
& {LoAd}	&\textbf{67.35 $\pm$ 0.97}	&\textbf{61.75 $\pm$ 0.54}	&\textbf{64.55} & 29.68 $\pm$ 1.52	& 31.43 $\pm$ 0.49& 30.56	\\ 
\hline
	& DANN	\cite{DANN}	& 57.94 $\pm$ 2.54 & 40.37 $\pm$ 0.53 	& 49.16 					& 22.71	$\pm$ 1.52	& 34.56	$\pm$ 0.36	&	28.63	\\
adapt on		& Auto-DIAL	\cite{carlucci2017auto}& 49.89 $\pm$ 0.19	& 44.78 $\pm$ 0.17	& 47.33							& 24.69 $\pm$ 0.10	& 25.22	$\pm$ 0.13	&	24.96	\\
sub-target	& {ROOTS}  \cite{roots}		& 54.29					& 55.29							& 54.79		& 20.07				& \textbf{34.71}			&	27.39	\\
& LoAd	& \textbf{67.10 $\pm$ 0.97	}	& \textbf{60.53 $\pm$ 0.53	}	&  	\textbf{63.82 }	& \textbf{28.38	$\pm$ 1.45	}		& 31.65	$\pm$ 0.49		& \textbf{30.01}		\\
\hline        
        \end{tabular}
	\end{center}
	\label{table:results}
    \vspace{-2mm}
\end{table*}

\subsection{Experimental Analysis} 
\label{subsec:analysis}
We evaluate the performance of our LoAd network in reducing the domain gap for the translation and scale settings described above.
As benchmark reference we use three domain adaptation approaches that have shown state of the art performance in 
computer vision for several non-robotic tasks:
\vspace{1mm}
\begin{list}{\labelitemi}{\leftmargin=1em}
\item{\textbf{DANN.}} The Domain Adversarial Neural Network \cite{DANN} takes as input both the labeled source and the
unlabeled target data and promotes the emergence of features that are discriminative for the main learning task
on the source domain and indiscriminate with respect to the shift between the domains. This is obtained by keeping
a single CNN path till the second fully connected layer and then doubling the final part of the network with the 
standard branch that minimizes the classification loss and a new branch that learns to confuse the domain discriminator.
\vspace{2mm}
\item{\textbf{Auto-DIAL.}} This is a deep learning network whose final objective is to 
minimize both the source classification loss and the target entropy loss by exploiting 
embedded domain alignment layers \cite{carlucci2017auto}. These layers perform batch normalization and induce 
a transformation of both the source and target distribution with a mixing
parameter which is learned automatically. 
\vspace{2mm}
\item{\textbf{ROOTS.}} We use this short name to indicate the first work about learning the spatial roots of visual 
domain shift \cite{roots}. As already mentioned in Section \ref{sec:related} this approach was based on a
computational intensive image patch occlusion process to produce domainness maps which were then statistically
evaluated to create image features at different domainness levels. Linear SVM classifiers are then trained on
the representations obtained at each level and the final annotation is based on the combined margin outputs. 
Here we adopt the strategy described in \cite{roots} to obtain the image representation, but we start from the maps 
produced through our efficient domain localization network (see section \ref{subsec:domloc}).
\end{list}
\vspace{1mm}
Finally, as non-adaptive reference we use standard {AlexNet}: the last fully connected layer of the network pre-trained on 
ImageNet is fine-tuned on the source samples and
tested on the target. 

Table \ref{table:results} reports the classification accuracy obtained by the reference methods listed above and by our LoAd network.
Let's focus first on the left part of the table containing 
results for the translation case.
Here both DANN and Auto-DIAL improve over the standard CNN with the second better than the first on average. As we 
know,  neither of these methods use local information but the image features are directly produced by end-to-end architectures.
On the other hand, ROOTS exploits the domainness maps but the AlexNet features extracted from different image patches 
are then hand-crafted. The result is a method that still overcomes standard CNN but has lower performance than the
other adaptive approaches. Finally our LoAd network outperforms all the competing methods. This remains true even
when the target data available during the training phase covers just a subset of the source categories (adapt on sub-target 
rows of the table). Here DANN and Auto-DIAL show a significant performance drop while the accuracy of LoAd
is almost unchanged showing its robustness. Note that ROOTS results also remain unchanged and this support the 
hypothesis that the  maps produced by the domain localization network when starting from a subset of 
categories contain the same information than that obtained over all the classes.

The right part of Table \ref{table:results} shows the classification accuracy on the scale domain case 
produced by all the considered methods.
As already indicated by the results in Table \ref{table:domain_shift},
object scaling induces a larger domain shift than the one caused by translation, thus the adaptation task is more challenging.
Here the best results are obtained by DANN, while LoAd improves on average over the AlexNet non adaptive baseline and over ROOTS but
underperforms with respect to DANN and Auto-DIAL. However, when considering only a subset of the target classes for
adaptation, the results obtained by DANN and Auto-DIAL decrease, while LoAd presents on average the best performance 
being just slightly affected by the change in training setting.

\begin{figure}
\begin{center}
\begin{tabular}{|c c|c | c|}
\hline
	\multicolumn{2}{|c|}{\multirow{2}{*}{images}}					  &  \multicolumn{2}{c|}{maps} \\
\cline{3-4}
						&	& 	left 	& right\\
\hline
{\multirow{2}{*}{left}}   & 
\includegraphics[width=0.07\textwidth]{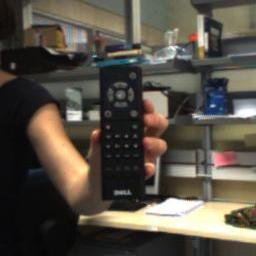}		&
\includegraphics[width=0.07\textwidth]{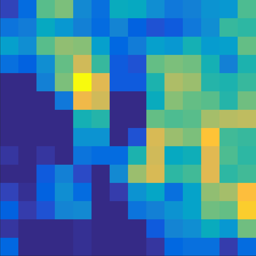}		&		
\includegraphics[width=0.07\textwidth]{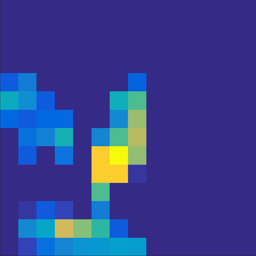}   \\
 &	
\includegraphics[width=0.07\textwidth]{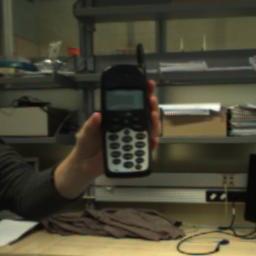}	&
\includegraphics[width=0.07\textwidth]{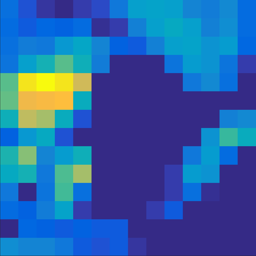}  &		
\includegraphics[width=0.07\textwidth]{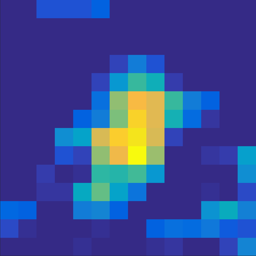} \\
\hline
{\multirow{2}{*}{right}} 	& 
\includegraphics[width=0.07\textwidth]{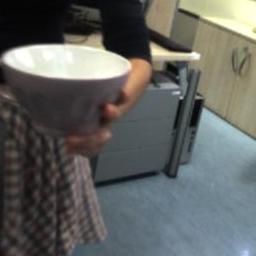} &	
\includegraphics[width=0.07\textwidth]{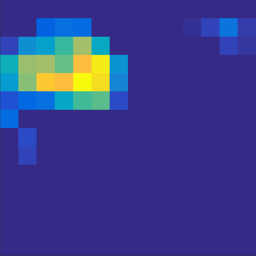}		&		
\includegraphics[width=0.07\textwidth]{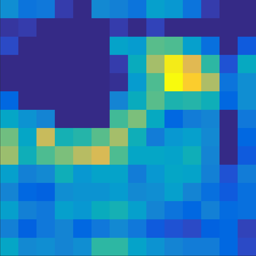} \\
 & 
\includegraphics[width=0.07\textwidth]{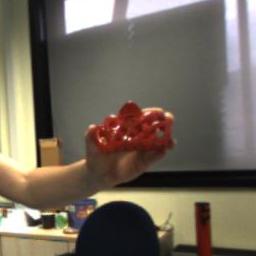} &	
\includegraphics[width=0.07\textwidth]{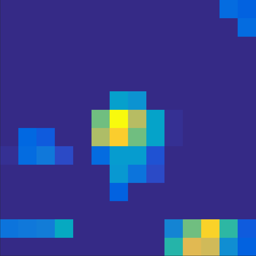}&			
\includegraphics[width=0.07\textwidth]{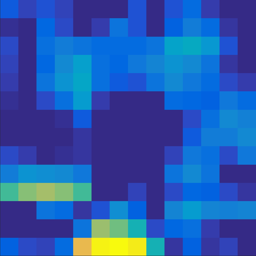}\\
\hline
\hline
							&		& 	close 	& far\\
\hline
{\multirow{2}{*}{close}} 	& 
\includegraphics[width=0.07\textwidth]{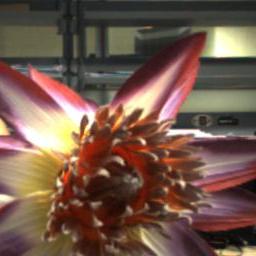}		&	
\includegraphics[width=0.07\textwidth]{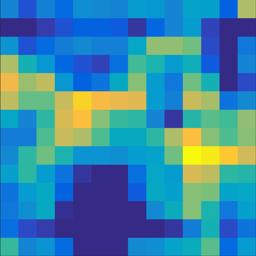}	&		
\includegraphics[width=0.07\textwidth]{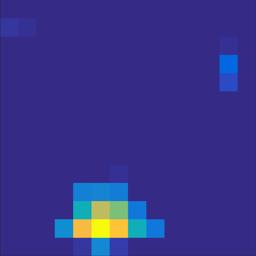} \\
							& 
\includegraphics[width=0.07\textwidth]{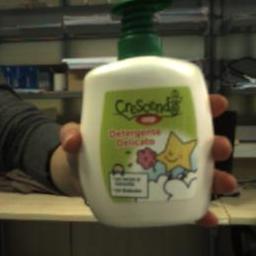} &
\includegraphics[width=0.07\textwidth]{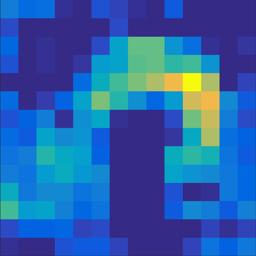} &
\includegraphics[width=0.07\textwidth]{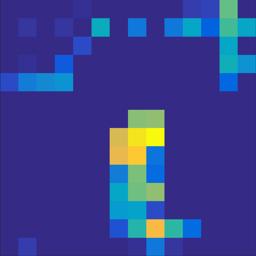} \\
\hline
{\multirow{2}{*}{far}} 		& 
\includegraphics[width=0.07\textwidth]{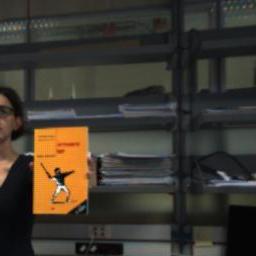}		&	
\includegraphics[width=0.07\textwidth]{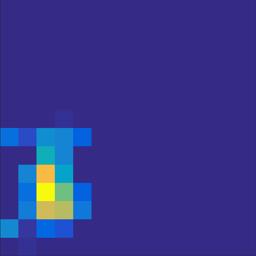} &		
\includegraphics[width=0.07\textwidth]{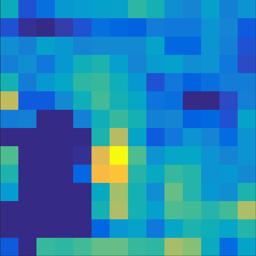} \\
							& 
\includegraphics[width=0.07\textwidth]{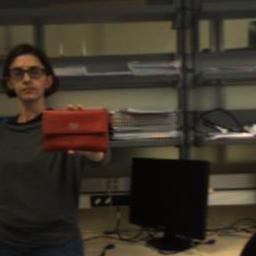}	&	
\includegraphics[width=0.07\textwidth]{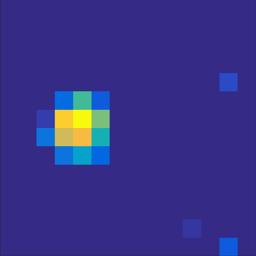}  	&		
\includegraphics[width=0.07\textwidth]{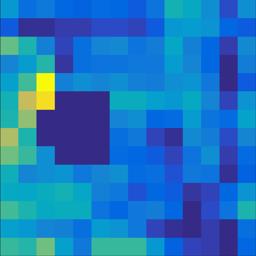} \\
\hline
\end{tabular}
\end{center}
\caption{Examples of domainness maps produced by the domain localization network: they highlight domain-specific 
and domain-generic image areas.}
\label{fig:domain_spec_gen}
\vspace{-1mm}
\end{figure}

\subsection{Discussion}
\label{subsec:ablation}
For a more in-depth analysis of the LoAd network, we present here an ablation study with
separate evaluations on its components.
\subsubsection{Domain localization performance}
\label{subsec:ablation:domainness}
The role and functionality of the domain localization network can be qualitatively analyzed by observing
the produced domainness maps. In the top part of Figure \ref{fig:domain_spec_gen} we show four
images from the translation setting, two from the \emph{left} domain and two from the \emph{right} domain. 
For each image the domain localization network produces two complementary maps: if the image belongs to 
the  \emph{left} domain, the left-map indicates what is domain-specific, while the right-map highlights
domain-generic areas \ie the image parts that mostly make this sample similar to the right domain. Analogously,
if the image belongs to the  \emph{right} domain, the left-map indicates domain-generic areas. By comparing
the maps with the images it can be easily seen that domain generic regions focus on the objects, while
the domain specific ones focus on the background which is typical for each domain. The LoAd network
uses the convolutional filters associated to the shown domain-generic maps while learning the object
classification model from the source data.

\begin{table}[tb!]
\caption{Effect of the careful design of the last CNN pooling layer which also allow a reduction in the number of neurons
in fully connected layers.} 
    \vspace{-4mm}
	\begin{center}
		\begin{tabular}{|c|c|c|c|} 
        \hline  
\multirow{2}{*}{} & \multicolumn{3}{c|}{iCubWorld Translation} \\
        \cline{2-4}
		             &  left $\rightarrow$ right 	& right $\rightarrow$ left 	& average\\
        \hline      
		pool5:$6\times6$, fc:4096      & 55.40 	& 57.40			& 56.40 \\
 		pool5:$4\times4$, fc:2048 		& 65.53		& 61.18			& 63.36	\\
		pool5:$3\times3$, fc:2048      & 65.23		& 60.87			& 63.05	\\
        \hline \hline
{\multirow{2}{*}{}}& \multicolumn{3}{c|}{iCubWorld Scale} \\
        \cline{2-4}
		&  close $\rightarrow$ far 	& far $\rightarrow$ close 	& average\\
        \hline      
		pool5:$6\times6$, fc:4096		& 17.91 	& 28.38			& 23.15 \\
		pool5:$4\times4$, fc:2048		& 25.64		& 29.49			& 27.57	\\
		pool5:$3\times3$, fc:2048		& 24.74		& 31.27			& 28.01	\\
        \hline 
        \end{tabular}
	\end{center}
	\label{table:pooling}
    \vspace{-4mm}
\end{table}

Similar considerations can be done for the scale case with examples shown in the bottom part of 
Figure \ref{fig:domain_spec_gen}. Here it can also be noticed that for the images of the \emph{close} 
domain the far-maps tend to highlight only a portion of the object as well as part of the top corners 
background which is actually shared among the domains. We speculate that this effect influences LoAd 
results on scale and can explain the lower performance with respect to the competing DANN in the 
whole-target adaptation setting. 

\subsubsection{Careful choice of spatial pooling dimensions}
\label{subsec:ablation:pooling}
When dealing with a domain shift that may be spatially grounded in the images, side tools that carefully manage 
localization can be extremely useful. The last pooling layer in standard deep architectures is at the interface 
between the convolutional part of the network that maintains local information from the images, and the fully 
connected part which instead disregards it. 
The design of this layer may have a significant impact on the final performance: we
experienced it by changing the final dimensionality of the feature maps and adaptively calculating the
window size and stride by following the same approach of \cite{sppooling}. The pool5 layer in AlexNet produces
maps of size $6\times6$ %
while we tested the effect of $4\times4$ and $3\times3$ maps.
Besides reducing the filter dimensionality, this choice allows also to decrease the number of neurons in the fully 
connected layers to $2048 \rightarrow 2048 \rightarrow K$ where $K=15$ in our setting. 

We report in Table \ref{table:pooling} the obtained results on one classification run of the translation and scale
domain shift tasks. Intuitively to reduce the map dimensionality we have to increase the pooling window size and this helps to produce 
a representation more invariant to the specific object location.
Our LoAd network was designed to include adaptive pooling layers producing $3\times3$ feature maps and fully connected
layers of 2048 dimensions. By comparing the results here with that reported in Table \ref{table:results}  we
can conclude that the careful design of the pooling layer has a relevant role in the final performance of LoAd.

\section{Conclusions}
\label{sec:concl}
We presented an approach to domain adaptation that explicitly leverages over the spatial roots of the domain shift between source 
and target domains in an end-to-end deep architecture. We showed how this is especially beneficial in robot vision, where the most 
common reasons for domain shift have indeed a strong spatial connotation, which leads to strong experimental results on robot vision 
data. A further advantage of our approach is its ability to handle scenarios where the classes in the target are a subset of the 
classes in the source domain at training time,
 a mostly unresearched scenario of great relevance in robot vision due to the challenges of accessing target data during training.
 
 Besides further testing of our approach, we will continue this work in three main directions: first, we will investigate up to 
 which point LoAd can be applied to the open set domain adaptation scenario depicted in \cite{openset2017}, with only a partial 
 overlap between the classes in the source and the target domain at training and \emph{test} time. Second, we will extend the 
 framework so to move from domain adaptation to domain generalization, i.e. the ability to perform well on any possible target 
 domain, as opposed to a given, specific one. Finally, we will move the approach to the multi-model setting, aiming for RGB-D 
 domain adaptation and generalization, as this scenario is of tremendous relevance for robotics applications.

\bibliographystyle{IEEEtran}
\bibliography{egbib}

\end{document}